\pdfoutput=1
\documentclass[11pt]{article}

\usepackage[]{acl}

\usepackage{times}
\usepackage{latexsym}
\usepackage{amsmath}
\usepackage{amsfonts,amssymb}
\usepackage{fancyhdr}
\usepackage{subfigure}
\usepackage{geometry}
\usepackage{newtxtext}
\usepackage{indentfirst}
\usepackage{array}
\usepackage{booktabs}
\usepackage{float}
\usepackage{bm}
\usepackage{color}
\usepackage{graphicx}
\usepackage{pythonhighlight}
\usepackage[title]{appendix}
\usepackage{booktabs}
\usepackage{multirow}
\usepackage{amssymb}
\usepackage{mathrsfs}
\usepackage{times}
\usepackage{latexsym}
\usepackage{algorithm}
\usepackage{algpseudocode}
\usepackage{enumitem}
\usepackage{color}
\usepackage{arydshln}
\usepackage{diagbox}

\usepackage[T1]{fontenc}
\usepackage[utf8]{inputenc}

\usepackage{microtype}
\usepackage{graphicx}

\makeatletter
\newcommand*\bigcdot{\mathpalette\bigcdot@{.5}}
\newcommand*\bigcdot@[2]{\mathbin{\vcenter{\hbox{\scalebox{#2}{$\m@th#1\bullet$}}}}}
\makeatother
\title{Missing Modality meets Meta Sampling (M$^3$S): An Efficient Universal Approach for Multimodal Sentiment Analysis with Missing Modality}

\author{Haozhe Chi \quad Minghua Yang \quad Junhao Zhu \quad Guanhong Wang \quad Gaoang Wang\thanks{\ \ Corresponding author.} \\
Zhejiang University-University of Illinois at Urbana-Champaign Institute, \\ Zhejiang University, China \\
\texttt{\{haozhe.20,\ minghua.20,\ junhao.20,\ gaoangwang\}@intl.zju.edu.cn} \\
\texttt{guanhongwang@zju.edu.cn}  \\
}

\begin{document}
\maketitle
\begin{abstract}
Multimodal sentiment analysis (MSA) is an important way of observing mental activities with the help of data captured from multiple modalities. However, due to the recording or transmission error, some modalities may include incomplete data. Most existing works that address missing modalities usually assume a particular modality is completely missing and seldom consider a mixture of missing across multiple modalities. In this paper, we propose a simple yet effective meta-sampling approach for multimodal sentiment analysis with missing modalities, namely Missing Modality-based Meta Sampling (M$^3$S). To be specific, M$^3$S formulates a missing modality sampling strategy into the modal agnostic meta-learning (MAML) framework. M$^3$S can be treated as an efficient add-on training component on existing models and significantly improve their performances on multimodal data with a mixture of missing modalities. We conduct experiments on IEMOCAP, SIMS and CMU-MOSI datasets, and superior performance is achieved compared with recent state-of-the-art methods.
\end{abstract}

\section{Introduction}
\noindent Multimodal sentiment analysis (MSA) aims to estimate human mental activities by multimodal data, such as a combination of audio, video, and text. Though much progress has been made recently, there still exist challenges, including missing modality problem. In reality, missing modality is a common problem due to the errors in data collection, storage, and transmission. To address the issue with missing modality in MSA, many approaches have been proposed \cite{ma2021smil,zhao2021missing,ma2021maximum,parthasarathy2020training,ma2021efficient,tran2017missing}.

% In recent years, missing modality problem has been attracting increasingly attention in multimodal study, especially in multimodal sentiment analysis. In multimodal learning, missing modality is common because it is often hard to get full modality in reality. To obtain better performance on missing modality data, several papers have proposed their methods \cite{ma2021smil}\cite{zhao2021missing}\cite{ma2021maximum}\cite{parthasarathy2020training}\cite{ma2021efficient}\cite{tran2017missing}.

In general, methods that address the missing modality issue usually only consider the situation where a certain input modality is severely damaged. The strategies of these proposed methods can be divided into three categories: 1) Designing new architectures with a reconstruction network to recover missing modality with the information from other modalities \cite{ma2021smil,ding2014latent}; 2) Formulating innovative and efficient loss functions to tackle missing modality \cite{ma2021efficient,ma2022multimodal}; 3) Improving the encoding and embedding strategies from existing models \cite{tran2017missing,cai2018deep}.
  
% In general, recent papers concerning missing modality mainly propose methods to handle situations when a certain input modality is severely damaged. There are mainly three strategies that have been proposed: (1) Design models with new architecture containing a  reconstruction network, which reconstructs missing modality by properly using input information from other modalities \cite{ma2021smil}\cite{ding2014latent}. (2) Design an innovate and efficient loss function or strategy to mathematically get a higher expectation in results' evaluation \cite{ma2021efficient}\cite{ma2022multimodal}. (3) Improve certain parts in existing model, including encoder and embedding ways \cite{tran2017missing}\cite{cai2018deep}.

\begin{figure}[t]
		\centering
        \includegraphics[width=0.95\linewidth]{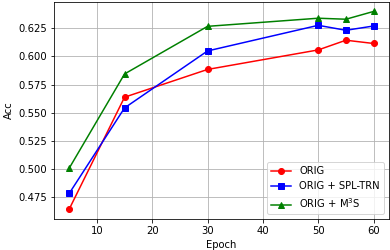}
		\caption{M$^3$S helps MMIN model achieve superior performance.}
		\label{MMIN_a}
\end{figure}

In the MSA tasks, most of the proposed methods focus on the situation where certain modalities are completely missing and the other modalities are complete. However, due to the transmission or collection errors, each modality may contain partial information based on a certain missing rate, while existing methods seldom consider this type of scenario and they are not suitable to be applied directly in this situation. Besides, our experiments also verify the inefficacy of existing methods in such a challenging situation, which is demonstrated in Section \ref{sec:analysis}. 

% However, in multimodal sentiment analysis task, while most of the proposed methods focus on data with a certain modality completely missing, few papers focus on the problem when each input modality has a certain missing rate. And no method has been proposed for data with medium missing rate in each modality specifically. Missing rate means percentage of missing data in total data. When facing this problem, we find that many previous methods could not apply. Also, some previous methods seem too complicated and could not cope with the situation efficiently.

To address the aforementioned problem, in this paper, we propose a simple yet effective solution to the \textbf{M}issing \textbf{M}odality problem with \textbf{M}eta \textbf{S}ampling in the MSA task, namely M$^3$S. To be specific, M$^3$S combines the augmented missing modality transform in sampling, following the model-agnostic meta-learning (MAML) framework \cite{finn2017model}. M$^3$S maintains the advantage of meta-learning and makes models easily adapt to data with different missing rates.
M$^3$S can be treated as an efficient add-on training component on existing models and significantly improve their performances on multimodal data with a mixture of missing modalities. We conduct experiments on IEMOCAP \cite{busso2008iemocap}, SIMS \cite{yu2020ch} and CMU-MOSI \cite{zadeh2016multimodal} datasets and superior performance is achieved compared with recent state-of-the-art (SOTA) methods. A simple example is shown in Figure~\ref{MMIN_a}, demonstrating the effectiveness of our proposed M$^3$S compared with other methods. More details are provided in the experiment section.

% To better address the problem, we propose M$^3$S, which is a simple but efficient method based on MAML training strategy in meta-learning \cite{finn2017model}. M$^3$S combines sampling method with meta-learning, which does data augmentation to input data. Furthermore, M$^3$S maintains the advantage of meta-learning, which makes model easily adapt to data with different missing rates. The whole framework is shown in Figure \ref{MMIN_a}.

The main contributions of our work are as follows:
\begin{itemize}%[leftmargin = 12pt]

\item We formulate a simple yet effective meta-training framework to address the problem of a mixture of partial missing modalities in the MSA tasks. 
\item The proposed method M$^3$S can be treated as an efficient add-on training component on existing models and significantly improve their performances on dealing with missing modality.
\item We conduct comprehensive experiments on widely used datasets in MSA, including IEMOCAP, SIMS, and CMU-MOSI. Superior performance is achieved compared with recent SOTA methods.

    % \item We notice a particular problem in multimodal sentiment analysis about medium missing modalities, which has not been studied before in missing modality field. This problem is of great significance because we often get incomplete modalities in reality. Moreover, we find that several outstanding models which handle multimodal sentiment analysis and missing modality well are showing bad performance with medium missing inputs. 
    % \item To handle this problem efficiently, we propose a simple but effective method called M$^3$S. Our method follows MAML strategy in meta-learning and can easily apply to different models in multimodal sentiment analysis. M$^3$S significantly improves model's performance on medium missing modalities and model's convergence rate. Also, M$^3$S adapts model to severely missing modalities. Thus M$^3$S can be an efficient way to handle medium missing modalities problems.
\end{itemize}

\section{Related Work}

% \noindent Emotion is the feeling produced by the interaction of life phenomena and human hearts. It involves subjective experience, physiological reaction, and behavioral reaction. Everyone has his own subjective feelings, and there are a series of physiological reactions. We express our emotions through facial expressions, words, body movements and other behavioral ways \cite{mao2021dialoguetrm}.\vspace{0.5em}

\subsection{Emotion Recognition}
\noindent {Emotion recognition} aims to identify and predict emotions through these physiological and behavioral responses. Emotions are expressed in a variety of modality forms. However, early studies on emotion recognition are often single modality. \citeauthor{shaheen2014emotion} \shortcite{shaheen2014emotion} and \citeauthor{calefato2017emotxt} \shortcite{calefato2017emotxt} present novel approaches to automatic emotion recognition from text. \citeauthor{burkert2015dexpression} \shortcite{burkert2015dexpression} and \citeauthor{deng2020multitask} \shortcite{deng2020multitask} conduct researches on facial expressions and the emotions behind them. \citeauthor{koolagudi2012emotion} \shortcite{koolagudi2012emotion} and \citeauthor{yoon2019speech} \shortcite{yoon2019speech} exploit acoustic data in different types of speeches for emotional recognition and classification tasks. Though much progress has been made for emotion recognition with single modality data, how to combine information from diverse modalities has become an interesting direction in this area. 
% \vspace{0.5em}

\subsection{Multimodal Sentiment Analysis}
\noindent {Multimodal sentiment analysis (MSA)} is a popular area of research in the present since the world we live in has several modality forms. When the dataset consists of more than one modality information, traditional single modality methods are difficult to deal with. MSA mainly focuses on three modalities: text, audio, and video. It makes use of the complementarity of multimodal information to improve the accuracy of emotion recognition. However, the heterogeneity of data and signals bring significant challenges because it creates distributional modality gaps. \citeauthor{hazarika2020misa} \shortcite{hazarika2020misa} propose a novel framework, MISA, which projects each modality to two distinct subspaces to aid the fusion process. And \citeauthor{hori2017attention} \shortcite{hori2017attention} introduce a multimodal attention model that can selectively utilize features from different modalities. Since the performance of a model highly depends on the quality of multimodal fusion, \citeauthor{han2021improving} \shortcite{han2021improving} construct a framework named MultiModal InfoMax (MMIM) to maximize the mutual information in unimodal input pairs as well as obtain information related to tasks through multimodal fusion process. Besides, \citeauthor{han2021bi} \shortcite{han2021bi} make use of an end-to-end network Bi-Bimodal Fusion Network (BBFN) to better utilize the dynamics of independence and correlation between modalities. Due to the unified multimodal annotation, previous methods are restricted in capturing differentiated information. \citeauthor{yu2021learning} \shortcite{yu2021learning} design a label generation module based on the self-supervised learning strategy. Then, joint training the multimodal and unimodal tasks to learn the consistency and difference. However, limited by the pre-processed features, the results show that the generated audio and vision labels are not significant enough.
\vspace{0.5em}

\subsection{Missing Modality Problem}
\noindent Compared with unimodal learning method, multimodal learning has achieved great success. It improves the performance of emotion recognition tasks by effectively combining the information from different modalities. However, the multimodal data may have missing modalities in reality due to a variety of reasons like signal transmission error and limited bandwidth. To deal with this problem, \citeauthor{ma2021maximum} \shortcite{ma2021maximum} propose an efficient approach based on maximum likelihood estimation to incorporate the knowledge in the modality-missing data. Nonetheless, the more complex scenarios like missing modalities exist in both training and testing phases are not involved. What's more, recent studies aim to capture the common information in different types of training data and leverage the relatedness among different modalities \cite{ma2021efficient,tran2017missing,parthasarathy2020training,wagner2011exploring}. To solve the problem that modalities will be missing is uncertain, \citeauthor{zhao2021missing} \shortcite{zhao2021missing} put forward a unified model: Missing Modality Imagination Network (MMIN). \citeauthor{ma2021smil} \shortcite{ma2021smil} utilize a new method named SMIL that leverages Bayesian meta-learning to handle the problem that modalities are partially severely missing, \textit{e.g.}, 90\% training examples may have incomplete modalities.

% The above-mentioned methods all focus on data with a certain modality completely missing or high missing rate. Our method is designed based on MAML training strategy in meta-learning \cite{finn2017model} to solve data with mixture of missing across modalities.

\section{Methodology}

\subsection{Problem Description}
\noindent The multimodal sentiment analysis aims at predicting the sentiment labels $\mathcal{Y}$ based on the model $f(\mathcal{X};\boldsymbol{\theta})$ given the multimodal data $\mathcal{X}$.
We consider the input data with three modalities, \textit{i.e.} $\mathcal{X}=(\mathcal{A},\mathcal{V},\mathcal{L})$, where $\mathcal{A}$, $\mathcal{V}$ and $\mathcal{L}$ represents audio, video and linguistic data, respectively. 
In this paper, we tackle the missing modality issue, where each modality can include missing data. 
% Denote the missing data generation for modality $\boldsymbol{M}$ as $\mathcal{G}(\boldsymbol{M}|\boldsymbol{\theta})$

% In multimodal sentiment analysis task, we note input data of three modalities as $\mathbf{(A,V,L)}$, standing for audio, video and linguistic data, and note missing rate as the percentage of missing data in total data. As described above, we focus on multimodal sentiment analysis task when three input modalities all have medium missing rate. That is, for each sample data $(A,V,L)$ (\textit{i.e.} $A$ belongs to $\mathbf{A}$, $V$ belongs to $\mathbf{V}$ and $L$ belongs to $\mathbf{L}$), missing rates of $A$,$V$,$L$ are all medium (\textit{i.e.}, within range $[0.4,0.6]$). 

\renewcommand{\algorithmicrequire}{\textbf{Input:}}
\renewcommand{\algorithmicensure}{\textbf{Output:}}
\begin{algorithm}[h]
	\caption{Meta-Sampling Training}
	\label{algorithm1}
	\begin{algorithmic}[1] 
	\Require Multimodal dataset $(\mathcal{X}=(\mathcal{A},\mathcal{V},\mathcal{L}),\mathcal{Y})$; number of iterations $K$ for inner loop; inner learning rate $\alpha$; 
	outer learning rate $\beta$; estimation model $f(\cdot \,;\boldsymbol{\theta})$; model's loss function ${l}\left(f,\mathcal{Y}\right)$.
% 	; model's loss function $f(A,V,L,\theta)$
	\While{not converged}
	\State Sample batch of data $\mathcal{X}_1$ and $\mathcal{X}_2$ from $\mathcal{X}$.
	\State Get $\Tilde{\mathcal{X}}_1=\mathcal{T}(\mathcal{X}_1\,; \mathcal{F})$ and $\Tilde{\mathcal{X}}_2=\mathcal{T}(\mathcal{X}_2\,; \mathcal{F})$.
    % \State Sample $(\Tilde{A}_1,\Tilde{V}_1,\Tilde{L}_1) \sim (X,Y,Z)$
    % \State Sample $(\Tilde{A}_2,\Tilde{V}_2,\Tilde{L}_2) \sim (X,Y,Z)$
    \State Set $\boldsymbol{\theta}_0 \gets \boldsymbol{\theta}$
    \State Meta-train:
    \For {$n=0$ to $K-1$}
    \State $\boldsymbol{\theta}_{n+1} \gets \boldsymbol{\theta}_n - \alpha\nabla_{\theta_n}\,
    l\left(f(\Tilde{\mathcal{X}}_1;\boldsymbol{\theta}_n), \mathcal{Y}_1\right)$
   \EndFor
   \State $\boldsymbol{\theta}^* \gets \boldsymbol{\theta}_K$
   \State Meta-update:
   \State $\boldsymbol{\theta} \gets \boldsymbol{\theta} - \beta\nabla_{\theta^*}\, {l}\left(f(\Tilde{\mathcal{X}}_2;\boldsymbol{\theta}^*), \mathcal{Y}_2\right)$
   \EndWhile
   \end{algorithmic} 
\end{algorithm} 

\subsection{Augmented Missing Modality Transform} \label{transform}

\noindent Given a sample $\boldsymbol{X}_i=(\boldsymbol{A}_i, \boldsymbol{V}_i, \boldsymbol{L}_i)$ from $\mathcal{X}$, we use a augmented transform $\mathcal{T}(\boldsymbol{X}_i\,; \mathcal{F})$ to generate a random sample with missing data based on a distribution $\mathcal{F}$. Specifically, for each modality $m \in \{a, v, l\}$, we define a missing ratio $r_m \in [0, 1]$, where $a$, $v$ and $l$ stands for audio, video and linguistic modality, respectively. For the encoded feature in each modality $m$, we replace the values between $[\lambda_m, \lambda_m+k_m-1]$ with zeros,
% (as shown in Figure \ref{missing_trans}), 
where $k_m$ represents the number of missing values with $k_m=\lfloor T_m\cdot r_m\rfloor$ and $T_m$ is the dimension of the encoded feature. $\lambda_m$ is sampled from the uniform distribution, \textit{i.e.}, $\lambda_m \sim \mathcal{U}(0, T_m-k_m)$. As a result, the augmented sample with missing modality can be obtained by $\Tilde{\boldsymbol{X}}_i = \mathcal{T}(\boldsymbol{X}_i\,; \mathcal{F})$, where $\mathcal{F}$ represents the composition of uniform distributions for each individual modality.

%     \begin{figure}[t]
% 		\centering
% % 		\setlength{\belowcaptionskip}{-1cm}
%         \includegraphics[width = 1.0\linewidth]{figures/step.jpg}
% 		\caption{Illustration of Missing Modality Transform}
% 		\label{missing_trans}
%     \end{figure}

\subsection{Training with Meta-Sampling}

\noindent Our M$^3$S follows MAML training framework \cite{finn2017model} with augmentation sampling. For each training iteration, we adopt the following steps.
% To train a certain model, we follow the process below in each training epoch:

First, we sample two independent batch of data, $\Tilde{\mathcal{X}}_1$ and $\Tilde{\mathcal{X}}_2$, based on the augmented missing modality transforms, $\mathcal{T}(\mathcal{X}_1\,; \mathcal{F})$ and $\mathcal{T}(\mathcal{X}_2\,; \mathcal{F})$, where the missing rate for each modality is determined by the sampling distribution $\mathcal{F}$. $\Tilde{\mathcal{X}}_1$ and $\Tilde{\mathcal{X}}_2$ are used as tasks from support set and query set, respectively, in the meta-learning.

% First we randomly select a missing rate for each modality $r_a, r_v, r_l$ from certain ranges. Then, two multimodal data $\boldsymbol{X}_1, \boldsymbol{X}_2$ are sampled from meta-sampling datasets $(X, Y, Z)$. Using the transfrom $\mathcal{T}(\boldsymbol{X}_i\,; \mathcal{F})$ mentioned above, we obtain the samples $\Tilde{\boldsymbol{X}}_1 = (\Tilde{A}_1,\Tilde{V}_1,\Tilde{L}_1), \Tilde{\boldsymbol{X}}_2 = (\Tilde{A}_2,\Tilde{V}_2,\Tilde{L}_2)$ with corresponding missing modality.

% $\Tilde{\boldsymbol{X}}_1$ is used as support set in inner loop training. 
Then, in the meta-train process, the model's parameter $\boldsymbol{\theta}$ is updated using gradient descent based on the loss function ${l}\left(f(\Tilde{\mathcal{X}}_1;\boldsymbol{\theta}), \mathcal{Y}_1\right)$ with the inner learning rate $\alpha$ for each iteration $n$ as follows:
\begin{equation}
    \boldsymbol{\theta}_{n+1} \gets \boldsymbol{\theta}_n - \alpha\nabla_{\theta_n}\, {l}\left(f(\Tilde{\mathcal{X}}_1;\boldsymbol{\theta}_n), \mathcal{Y}_1\right),
\end{equation}
where $\mathcal{Y}_1$ is the set of sentiment labels of $\Tilde{\mathcal{X}}_1$, and the loss function ${l}\left(f(\Tilde{\mathcal{X}}_1;\boldsymbol{\theta}), \mathcal{Y}_1\right)$ is determined by loss used in each base model (\textit{i.e.}, MMIM, MISA, Self-MM, MMIN. See Section \ref{sec:base model} for more details).
The meta-train process is conducted for $K$ iterations. We denote $\boldsymbol{\theta}_{K}$ as $\boldsymbol{\theta}^*$. 

\begin{figure*}[t]
		\centering
        \includegraphics[width = 1.0\linewidth]{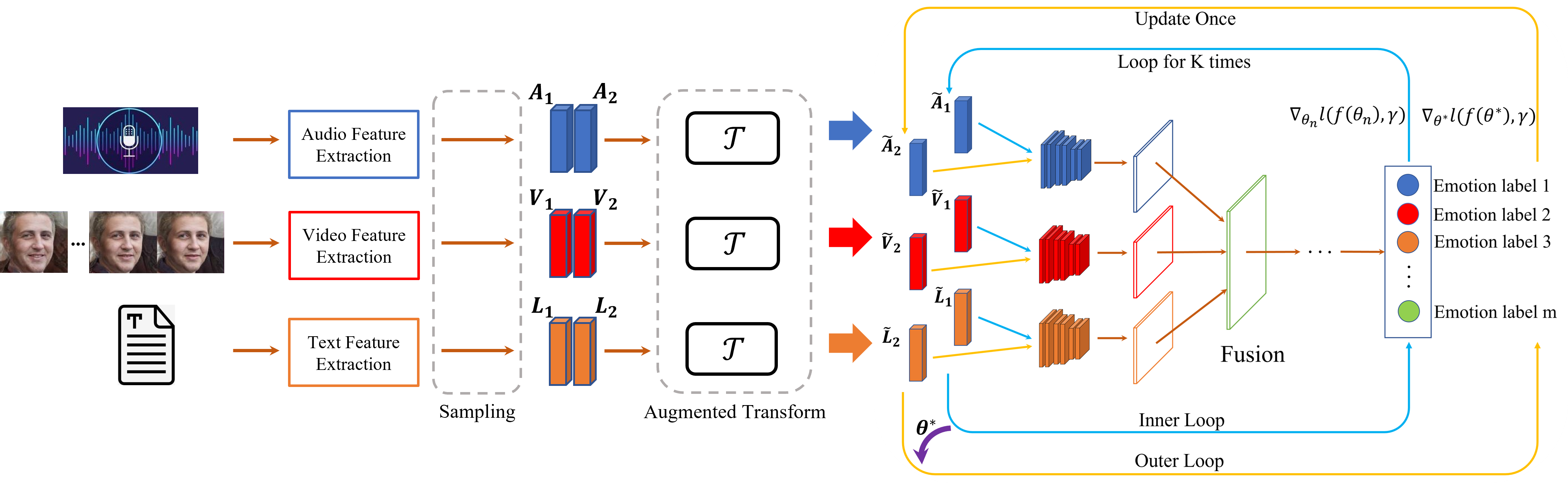}
		\caption{The Overall Architecture of M$^3$S. We first use augmented transform to generate two batches of data for features from each modality. Then the meta-train and meta-update are conducted on the two batches of data to learn the model parameters $\boldsymbol{\theta}$.}
		\label{meta_sampling}
\end{figure*}

% When updating is finished, the final parameter vector $\theta_K$ is taken as $\theta^*$.
Finally, we use the query set $\Tilde{\mathcal{X}}_2$ and its set of sentiment labels $\mathcal{Y}_2$ in the outer loop meta-update step. The model parameters are updated with the learning rate $\beta$ as follows:
% It is then used to perform the outer loop meta-update process with query set $\Tilde{\boldsymbol{X}}_2$ and outer learning rate $\beta$, such that the network parameters $\theta$ are updated as follows:
\begin{equation}
    \boldsymbol{\theta} \gets \boldsymbol{\theta} - \beta\nabla_{\theta^*}\, {l}\left(f(\Tilde{\mathcal{X}}_2;\boldsymbol{\theta}^*), \mathcal{Y}_2\right).
\end{equation}
The whole algorithm in general case is shown in Algorithm \ref{algorithm1} and Figure \ref{meta_sampling} illustrates the meta-sampling training process.

\section{Experiment Setup}
\noindent In this section, we present the setup of our experiments, including the used datasets, baseline methods, evaluation metrics, and implementation details of the proposed method. 

\subsection{Datasets}
\label{section:datasets}
\noindent We conduct our experiments on the following three datasets, \textit{i.e.}, IEMOCAP \cite{busso2008iemocap}, SIMS \cite{yu2020ch} and CMU-MOSI \cite{zadeh2016multimodal}. The statistics of the datasets are reported in Table~\ref{tab:data}.
\begin{itemize}%[leftmargin = 9pt]

\item \textbf{IEMOCAP} comprises of several recorded videos in 5 conversation sessions, and each session contains many scripted plays and dialogues. The actors performed selected emotional scripts and also improvised hypothetical scenarios designed to elicit specific types of emotions, which provided detailed information about their facial expressions and hand movements.
% We apply M$^3$S to baseline model MMIN and use the extracted features of IEMOCAP \cite{zhao2021missing} to conduct experiments on MMIN.

\item \textbf{SIMS} dataset is a multimodal sentiment analysis benchmark containing 2281 video clips from various sources (\textit{i.e.}, movies, shows, TV serials, etc.). SIMS contains fine-grained annotations of different modalities and includes people's natural expressions in video clips. And each sample in SIMS dataset is labeled with a score from -1 to 1, standing for sentiment response (\textit{i.e.}, from strongly negative to strongly positive).

\item \textbf{CMU-MOSI} has 2199 video segments in total, which are sliced from 93 YouTube videos. The videos address a large array of topics like books, products, and movies. In these video segments, 89 narrators show their opinions on different topics. Most of the speakers are around 20-30 years old. They all express themselves in English, although they come from different countries. 
% We apply M$^3$S to baseline models MMIM and MISA and conduct experiments on these two models on CMU-MOSI dataset.
\end{itemize}

\begin{table}[!t]
    \centering
    % \resizebox{\linewidth}{13mm}{
    \begin{tabular}{lcccc}
    \toprule
    Dataset & Train & Valid & Test & All \\
    % \cmidrule(r){1-1} \cmidrule(r){2-5}
    \midrule
    % \specialrule{0em}{2pt}{2pt}
    SIMS & 1368 & 456 & 457 & 2281 \\
    % \specialrule{0em}{2pt}{2pt}
    MOSI & 1284 & 229 & 686 & 2199 \\
    % \specialrule{0em}{2pt}{2pt}
    IEMOCAP & 4446 & 3342 & 3168 & 10956  \\
    % \specialrule{0em}{2pt}{2pt}
    \bottomrule%[1.2pt]
    \end{tabular}
    \caption{Statistics of the Used Datasets}
    \label{tab:data}
\end{table}

\begin{table*}[t]
\centering
\resizebox{1.0\linewidth}{!}{
\begin{tabular}{cccccccc}
  \toprule
  \multirow{2}{*}{Method} & \multicolumn{4}{c}{Self-MM (SIMS)} & \multicolumn{3}{c}{MMIN (IEMOCAP)}  \\
  \cmidrule(r){2-5} \cmidrule(r){6-8}
      & MAE & Corr & Acc-2 & F1-Score & Acc & Uar & F1-Score  \\
  \hline
  %\specialrule{0em}{2pt}{2pt}
  ORIG & 0.5171  & 0.3918   & 0.7291   & 0.6980   & 0.6136   & 0.6403   & 0.6049  \\
  %\specialrule{0em}{5pt}{5pt}
  ORIG + SPL-TRN & \textbf{0.5049}  & 0.4080  & 0.7392  & 0.7102  & 0.6357  & 0.6518  & 0.6235   \\
  %\specialrule{0em}{5pt}{5pt}
  \hdashline
  \specialrule{0em}{1pt}{1pt}
  ORIG + M$^3$S & 0.5053   &\textbf{0.4091}   &\textbf{0.7405}   &\textbf{0.7119}   &\textbf{0.6398}   &\textbf{0.6536}   &\textbf{0.6296}  \\
  $\Delta_{ORIG}$ & \textcolor[rgb]{0,0.5,0}{$\downarrow0.0118$} & \textcolor[rgb]{0,0.5,0}{$\uparrow0.0173$} & \textcolor[rgb]{0,0.5,0}{$\uparrow0.0114$} & \textcolor[rgb]{0,0.5,0}{$\uparrow0.0139$} & 
  \textcolor[rgb]{0,0.5,0}{$\uparrow0.0262$} & 
  \textcolor[rgb]{0,0.5,0}{$\uparrow0.0133$} &
  \textcolor[rgb]{0,0.5,0}{$\uparrow0.0247$}   \\
  \hline\hline
  \specialrule{0em}{1pt}{1pt}
  \multirow{2}{*}{Method} & \multicolumn{3}{c}{MISA (MOSI)} & - & \multicolumn{3}{c}{MMIM (MOSI)}  \\
  \cmidrule(r){2-4} \cmidrule(r){6-8}
   & MAE & Corr & Acc-7 & - & MAE & Corr & Acc-7 \\
  \hline
  ORIG & 0.8886   & 0.7349   & 0.3863 & - & 0.7175  & 0.7883   & 0.4592  \\
  ORIG + SPL-TRN & \textbf{0.8279}  & \textbf{0.7355} & 0.4155 & - & 0.7126  & 0.7825 & 0.4650 \\
  \hdashline
  \specialrule{0em}{1pt}{1pt}
  ORIG + M$^3$S & 0.8393  & 0.7346   &\textbf{0.4282} & - &\textbf{0.7014}   &\textbf{0.7985}   &\textbf{0.4852}  \\
  $\Delta_{ORIG}$ & \textcolor[rgb]{0,0.5,0}{$\downarrow0.0493$} & \textcolor[rgb]{0,0,0}{$\downarrow0.0003$} & \textcolor[rgb]{0,0.5,0}{$\uparrow0.0419$} & - & 
  \textcolor[rgb]{0,0.5,0}{$\downarrow0.0161$} & 
  \textcolor[rgb]{0,0.5,0}{$\uparrow0.0102$} &
  \textcolor[rgb]{0,0.5,0}{$\uparrow0.0260$}   \\
  %\specialrule{0em}{2pt}{2pt}
  \bottomrule
\end{tabular}}
\caption{Results of four baseline models with different training methods applied. Input and test data both have missing rates between 40\% and 60\%. ORIG stands for original model; SPL-TRN stands for sampling-training. $\Delta_{ORIG}$ presents the improved performance based on original model that M$^3$S has achieved.}
\label{results}
% \vspace{-10pt}
\end{table*}

\subsection{Baseline Methods}
\label{sec:base model}

\noindent We use four recent SOTA methods for comparison in the experiments. The methods include MMIM \cite{han2021improving}, MISA \cite{hazarika2020misa}, Self-MM \cite{yu2021learning} and MMIN \cite{zhao2021missing}, which are summarized as follows. 
\begin{itemize}
\item[\dag] \textbf{MMIM} helps mutual information reach maximum and maintains information related to tasks during the process of multimodal fusion, which shows significant results in multimodal sentiment analysis tasks.
\item[\dag] \textbf{MISA} is a novel model in emotion recognition that represents modality more effectively and improves the fusion process significantly.
\item[\dag] \textbf{Self-MM} has novel architecture containing several innovative modules (like a module for label generation) and reaches brilliant results in multimodal sentiment analysis tasks.
\item[\dag] \textbf{MMIN} handles the problem that input data has uncertain modalities completely missing and achieves superior results under various missing modality conditions.
\end{itemize}

\subsection{Evaluation Metrics}
\noindent 
Following the four baseline methods mentioned above, we use the following evaluation metrics, including mean absolute error (MAE), Pearson correlation (Corr), binary classification accuracy (Acc-2), weighted F1 score (F1-Score), accuracy score (Acc), unweighted average recall (Uar), and seven-class classification accuracy (Acc-7). Acc-7 denotes the ratio of predictions that are in the correct interval among the seven intervals ranging from -3 to 3. For all metrics, higher values show better performance except for MAE.

\subsection{Implementation Details} 
\label{Details}

\paragraph{Hyperparameter Settings.} The settings of inner learning rate, outer learning rate and batch size $\{\alpha$, $\beta$, batch\_size$\}$ are as follows: MMIN $\{$2e-4, 1e-4, 256$\}$; MMIM $\{$1e-3, 1e-3, 32$\}$; MISA $\{$1e-4, 1e-4, 128$\}$; For Self-MM, the learning rate for three modalities $\{\mathcal{A},\mathcal{V},\mathcal{L}\}$ is $\{$5e-3, 5e-3, 5e-5$\}$, and the batch size is 32. 

\paragraph{Feature Extraction Details.} Following the baseline methods, we adopt the extracted features as the input for each modality. The feature extraction methods on each modality $\{\mathcal{A},\mathcal{V},\mathcal{L}\}$ are listed as follows: MMIN $\{$OpenSMILE-"IS13\_ComParE" \cite{eyben2010opensmile}, DenseNet \cite{huang2017densely} trained on FER+ corpus \cite{barsoum2016training}, BERT \cite{devlin2018bert}$\}$; Self-MM, MMIM, MISA $\{$sLSTM \cite{hochreiter1997long}, sLSTM, BERT$\}$.

\paragraph{Experimental Details.} We use Adam as the optimizer for all four baseline models. The training epoch for $\{$MMIN, MMIM, MISA$\}$ is $\{60, 40, 500\}$. Self-MM adopts the "early stop" strategy to obtain the best result. Therefore, its training epoch is unfixed. In Section \ref{main_results}, We compare the performance of three different training methods dealing with missing modalities in our experiment results: 1) original model's training method (ORIG), where the missing rate of each sample is fixed along the training process during different epochs; 2) original model with Sampling-Training strategy applied (ORIG + SPL-TRN), which adopts augmented sampling without meta-learning process, as illustrated in Section \ref{transform}; 3) original model with M$^3$S added on (ORIG + M$^3$S), which is the proposed method.

\section{Results and Analysis}
\label{sec:analysis}

% \noindent In this section, we show our results and ablation study. Also, we further discuss about our results and give detailed analyses.

% \subsection{Comparative Analysis}
% \noindent Comparison with state-of-the-art (SOTA) methods

\subsection{Main Results}
\label{main_results}

\begin{table*}[t]
    \centering
    \resizebox{1.0\linewidth}{!}{
    \begin{tabular}{cccccccc}
    \toprule
    \multirow{2}{*}{Input Missing Rate} & \multirow{2}{*}{Method} & \multicolumn{3}{c}{MMIN (IEMOCAP)} & \multicolumn{3}{c}{MMIM (MOSI)}  \\
    \cmidrule(r){3-5} \cmidrule(r){6-8}
      & & Acc & Uar & F1-Score & MAE & Corr & Acc-7  \\
    \midrule
    %\specialrule{0em}{1pt}{1pt}
    \multirow{4}{*}{60\% $\sim$ 80\%} & ORIG & 0.5849 & 0.5915 & 0.5748 & \textbf{0.7132} & \textbf{0.7905} & 0.4577  \\
    %\specialrule{0em}{1pt}{1pt}
     & ORIG + SPL-TRN & 0.5812 & 0.5901 & 0.5689 & 0.7268 & 0.7867 & 0.4549  \\
    %\specialrule{0em}{1pt}{1pt}
    \cdashline{2-8}
    \specialrule{0em}{1pt}{1pt}
     & ORIG + M$^3$S & \textbf{0.5900} & \textbf{0.6026} & \textbf{0.5764} & 0.7208 & 0.7890 & \textbf{0.4588}  \\
     & $\Delta_{ORIG}$ & \textcolor[rgb]{0,0.5,0}{$\uparrow0.0051$} & \textcolor[rgb]{0,0.5,0}{$\uparrow0.0111$} & \textcolor[rgb]{0,0.5,0}{$\uparrow0.0016$} & 
    \textcolor[rgb]{0,0,0}{$\uparrow0.0076$} & 
    \textcolor[rgb]{0,0,0}{$\downarrow0.0015$} &
    \textcolor[rgb]{0,0.5,0}{$\uparrow0.0011$}   \\
    %\specialrule{0em}{1pt}{1pt}
    \midrule
    \multirow{4}{*}{40\% $\sim$ 60\%} & ORIG & 0.6136   & 0.6403   & 0.6049 & 0.7175  & 0.7883   & 0.4592   \\
     & ORIG + SPL-TRN & 0.6357  & 0.6518  & 0.6235 & 0.7126  & 0.7825 & 0.4650   \\
    \cdashline{2-8}
    \specialrule{0em}{1pt}{1pt}
     & ORIG + M$^3$S &\textbf{0.6398}  &\textbf{0.6536}  &\textbf{0.6296} &\textbf{0.7014}   &\textbf{0.7985}   &\textbf{0.4852}   \\
      & $\Delta_{ORIG}$ & \textcolor[rgb]{0,0.5,0}{$\uparrow0.0262$} & \textcolor[rgb]{0,0.5,0}{$\uparrow0.0133$} & \textcolor[rgb]{0,0.5,0}{$\uparrow0.0247$} & 
    \textcolor[rgb]{0,0.5,0}{$\downarrow0.0161$} & 
    \textcolor[rgb]{0,0.5,0}{$\uparrow0.0102$} &
    \textcolor[rgb]{0,0.5,0}{$\uparrow0.0260$}   \\
    \midrule
    %\specialrule{0em}{1pt}{1pt}
    \multirow{4}{*}{20\% $\sim$ 40\%} & ORIG & 0.6192 & 0.6453 & 0.6078 & 0.7129 & 0.7893 & 0.4694 \\
    %\specialrule{0em}{1pt}{1pt}
     & ORIG + SPL-TRN & 0.6335 & \textbf{0.6513} & 0.6221 & 0.7218 & 0.7832 & 0.4665  \\
    %\specialrule{0em}{1pt}{1pt}
    \cdashline{2-8}
    \specialrule{0em}{1pt}{1pt}
     & ORIG + M$^3$S & \textbf{0.6367} & 0.6504 & \textbf{0.6266} & \textbf{ 0.7049} & \textbf{0.7923} & \textbf{0.4838}  \\
      & $\Delta_{ORIG}$ & \textcolor[rgb]{0,0.55,0}{$\uparrow0.0175$} & \textcolor[rgb]{0,0.5,0}{$\uparrow0.0051$} & \textcolor[rgb]{0,0.5,0}{$\uparrow0.0188$} & 
    \textcolor[rgb]{0,0.5,0}{$\downarrow0.0080$} & 
    \textcolor[rgb]{0,0.5,0}{$\uparrow0.0030$} &
    \textcolor[rgb]{0,0.5,0}{$\uparrow0.0144$}   \\
    %\specialrule{0em}{1pt}{1pt}
    \bottomrule
    \end{tabular}}
    \caption{Results on MMIN and MMIM under three different missing rate levels. Test data have the same range of missing rates as input data.}
    \label{miss_rate}
\end{table*}

% \begin{table*}[t]
%     \centering
%     % \resizebox{\linewidth}{18mm}{
%     \begin{tabular}{ccccccc}
%     \toprule[1.3pt]
%     \multirow{2}{*}{\textbf{\large Method}} & \multicolumn{3}{c}{\textbf{MMIN (IEMOCAP)}} & \multicolumn{3}{c}{\textbf{MMIM (MOSI)}}  \\
%     \cmidrule(r){2-4} \cmidrule(r){5-7}
%       & ACC & Uar & F1-Score & MAE & Corr & $Mult_{Acc-7}$  \\
%     \midrule[1.1pt]
%     \specialrule{0em}{2pt}{2pt}
%     \textbf{ORIG} & 0.6192 & 0.6453 & 0.6078 & 0.7129 & 0.7893 & 0.4694 \\
%     \specialrule{0em}{3pt}{3pt}
%     \textbf{ORIG + SPL-TRN} & 0.6335 & 0.6513 & 0.6221 & 0.7218 & 0.7832 & 0.4665  \\
%     \specialrule{0em}{3pt}{3pt}
%     \textbf{ORIG + M$^3$S} & \textbf{0.6367} & 0.6504 & \textbf{0.6266} & \textbf{ 0.7049} & \textbf{0.7923} & \textbf{0.4738}  \\
%     \specialrule{0em}{1pt}{1pt}
%     \bottomrule[1.3pt]
%     \end{tabular}
%     \caption{Results on MMIN and MMIM. ORIG stands for original model; SPL-TRN stands for sampling-training; The inputs all have missing rate $\leq$ 30\%.}
%     \label{}
% \end{table*}

\begin{figure*}[t]
		\centering
		\subfigure[Valid Loss]{
			\includegraphics[width = 0.47\linewidth]{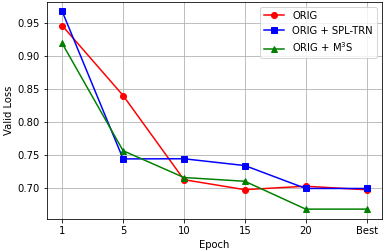}
			\label{valid_loss}
		}
        \quad
		\subfigure[Test Loss]{
			\includegraphics[width = 0.47\linewidth]{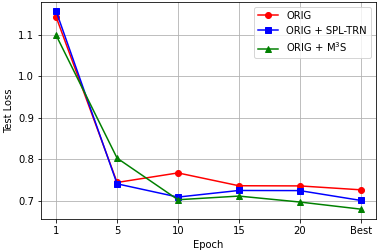}
			\label{test_loss}
		}
		\caption{Validation and testing losses of three methods along training built on the MMIM Model.} %Red, blue and green colors represent ORIG, ORIG+SPL-TRN, and ORIG+M$^3$S, respectively.}
		%\label{}
\end{figure*}

\noindent Built on the baseline models, we conduct experiments with the proposed M$^3$S method and show its effectiveness in Table \ref{results}. The missing rate is set as the medium rate, between 40\% and 60\%. Since M$^3$S can be an add-on component to existing methods with the capability of dealing with missing modality, we compare M$^3$S with Sampling-Training (SPL-TRN) and four original baseline methods. For all the testing datasets, M$^3$S achieves superior performance in almost all evaluation metrics compared with the original baseline methods, as expected. Since SPL-TRN only adopts augmented sampling without meta-learning process, it achieves worse performance than our M$^3$S method in most of the experiments. This result demonstrates that the meta-sampling training process can better learn the common knowledge from other modalities to deal with the missing information. It also verifies that meta-training can better utilize the information from random augmentations.
As a matter of fact, with the help of M$^3$S, MMIN model achieves the highest Acc, highest Uar, and highest F1-Score. Also, built upon the other three baselines (Self-MM, MISA, MMIM), M$^3$S helps in reaching the lowest MAE, highest Corr, and highest Acc in most situations, which shows the efficiency and universality of M$^3$S.

% In order to better show the effectiveness of M$^3$S, we conduct experiments on four baseline models in multimodal sentiment analysis. As shown in Table \ref{results}, we set the inputs' missing rate between 40\% and 60\%. When sampling-training method is applied to four models, models achieve better performance compared to those just trained with their original process. Sampling method means we do not follow MAML strategy in training and directly sample data from meta-sampling datasets (described in Section \ref{Details}: Details for Performing M$^3$S) to train models. This would enhance input data and help model achieve better results. And when M$^3$S method is applied, four models reach superior results in almost all indexes in Table \ref{results}. This shows that sampling training combined with MAML strategy does stronger data augmentation and is more effective in improving models' performance.

\subsection{Studies of Various Missing Rates}

\noindent To verify the effectiveness of methods on different missing rates, we conduct experiments on two datasets by varying the input missing rate to three levels (\textit{i.e.}, 20\%-40\%, 40\%-60\%, and 60\%-80\%). Results in Table \ref{miss_rate} show that for nearly all the cases, our method M$^3$S outperforms ORIG and ORIG+SPL-TRN methods. Specifically, when input missing rate falls within the range 40\%-60\%, ORIG+M$^3$S shows the greatest increment in all metrics, which shows that M$^3$S achieves the most significant effect on models with medium missing level.

% \multirow{3}{*}{60\% $\sim$ 80\%} & ORIG & 0.6035 & 0.6281 & 0.5953 & 0.7201 & 0.7794 & 0.4534  \\
% %\specialrule{0em}{1pt}{1pt}
%  & ORIG + SPL-TRN & 0.6152 & 0.6166 & 0.6023 & 0.7412 & 0.7695 & 0.4461  \\
% %\specialrule{0em}{1pt}{1pt}
%  & ORIG + M$^3$S & \textbf{0.6206} & 0.6140 & \textbf{0.6072} & \textbf{0.7025} & \textbf{0.7884} & \textbf{0.4825}

% \noindent To do ablation study on inputs with different missing rates, we further conduct experiments on two baselines with severely and mildly missing inputs. The same as what is shown in Table \ref{results}, Table \ref{miss_rate} reveals that M$^3$S helps two models achieve superior evaluation metrics. This comparative result also shows that our method helps models adapt to inputs with large as well as small missing rate.

\subsection{Convergence Comparison}

\begin{figure}[t]
		\centering
		\subfigure[Uar]{
			\includegraphics[width = 0.95\linewidth]{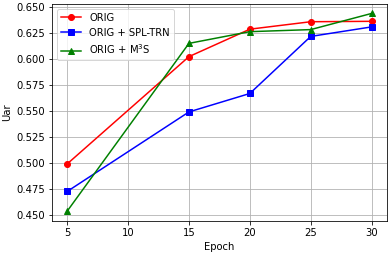}
			\label{uar}	
		}
        % \quad
		\subfigure[F1-Score]{
			\includegraphics[width = 0.95\linewidth]{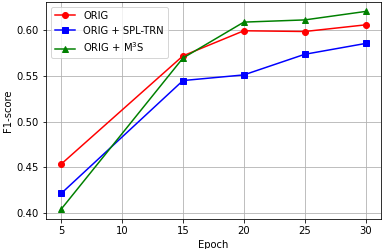}
			\label{f1}	
		}
		\caption{Uar and F1-Score of three methods along training built on the MMIN Model.} %Red, blue and green colors represent ORIG, ORIG+SPL-TRN, and ORIG+M$^3$S, respectively.}
\end{figure}

% \noindent As shown in Table \ref{results}, we conduct experiments on four baselines by applying different training method to each model and get comparative results. The results show that our method significantly improves nearly all evaluation metrics and thus proves its universality and effectiveness. Moreover, as shown in Table \ref{miss_rate}, we conduct experiments on two baselines with severely missing inputs. The comparative results in table show that our method helps models adapt to inputs with large missing rate and achieve superior evaluation metrics.

\noindent As is shown in Figure \ref{valid_loss} and \ref{test_loss}, we plot the process of MMIM model's loss decline. It is clearly shown in plots that M$^3$S helps original model converge to the lowest loss after 10 to 15 epochs of training. As shown in Figure \ref{uar} and Figure \ref{f1}, we also select MMIN model and plot its convergence process because the trend of its metrics changes more obviously. These two figures, along with Figure \ref{MMIN_a} show the characteristic of our method: although M$^3$S does not show strong competitiveness in the first few epochs, with the progress of training, M$^3$S helps model achieve faster growth of various metrics and finally converge to a higher result. 

\begin{table}[t]
\begin{minipage}[c]{1.0\linewidth}
    \centering
    \resizebox{1.0\linewidth}{!}{
    \begin{tabular}{c|c:c:c:c}
    \hline
    \multirow{2}{*}{MMIN} & \multirow{2}{*}{ORIG} & ORIG + & ORIG + & \multirow{2}{*}{\large$\Delta_{ORIG}$}  \\
     & & SPL-TRN & M$^3$S &  \\
    \hline
    Acc & 0.6035 & 0.6152 & \textbf{0.6206} & \textcolor[rgb]{0,0.5,0}{$\uparrow0.0171$}  \\%[0.5ex]
    Uar & \textbf{0.6281} & 0.6166 & 0.6140 & \textcolor[rgb]{0,0,0}{$\downarrow0.0141$}  \\%[0.5ex]
    F1-Score & 0.5953 & 0.6023 & \textbf{0.6072} & \textcolor[rgb]{0,0.5,0}{$\uparrow0.0119$} \\
    \hline
    \end{tabular}}
\end{minipage}
\begin{minipage}[c]{1.0\linewidth}
    \centering
    \resizebox{1.0\linewidth}{!}{
    \begin{tabular}{c|c:c:c:c}
    \hline
    \multirow{2}{*}{MMIM} & \multirow{2}{*}{ORIG} & ORIG + & ORIG + & \multirow{2}{*}{\large$\Delta_{ORIG}$}  \\
     & & SPL-TRN & M$^3$S &  \\
    \hline
    MAE & 0.7201 & 0.7412 & \textbf{0.7025} & \textcolor[rgb]{0,0.5,0}{$\downarrow0.0176$}  \\%[0.5ex]
    Corr & 0.7794 & 0.7695 & \textbf{0.7884} & \textcolor[rgb]{0,0.5,0}{$\uparrow0.0090$}  \\%[0.5ex]
    \ \ \ Acc-7\ \ \  & 0.4534 & 0.4461 & \textbf{0.4825} & \textcolor[rgb]{0,0.5,0}{$\uparrow0.0291$} \\
    \hline
    \end{tabular}}
\end{minipage}
    \caption{Results on MMIN (IEMOCAP) and MMIM (MOSI), where input data have missing rates 40\%-60\% and test data have missing rates 60\%-80\%.}
    \label{adapt_severe}
\end{table}

\begin{table*}[t]
\centering
\resizebox{1.0\linewidth}{!}{
\setlength{\tabcolsep}{3mm}{
\begin{tabular}{cccccccc}
  \toprule
  \multirow{2}{*}{P-value of t-test} & \multicolumn{4}{c}{Self-MM (SIMS)} & \multicolumn{3}{c}{MMIN (IEMOCAP)}  \\
  \cmidrule(r){2-5} \cmidrule(r){6-8}
      & MAE & Corr & Acc-2 & F1-Score & Acc & Uar & F1-Score  \\
  \hline
  \specialrule{0em}{2pt}{2pt}
  $P(T\leq t)$ & 0.1959  & 0.0384  & 0.0018  & 0.0615  & 0.0007 & 7.95E-5 & 0.0005  \\
  \specialrule{0em}{2pt}{2pt}
  \hline\hline
  \specialrule{0em}{1pt}{1pt}
  \multirow{2}{*}{P-value of t-test} & \multicolumn{3}{c}{MISA (MOSI)} & - & \multicolumn{3}{c}{MMIM (MOSI)}  \\
  \cmidrule(r){2-4} \cmidrule(r){6-8}
   & MAE & Corr & Acc-7 & - & MAE & Corr & Acc-7 \\
  \hline
  \specialrule{0em}{2pt}{2pt}
  $P(T\leq t)$ & 0.0473  & 0.1873  & 0.0405 & - & 0.0277  & 0.1971  & 0.0263  \\
  \bottomrule
\end{tabular}}}
\caption{Two-tailed significance test (t-test) of M$^3$S.}
\label{significance_test}
% \vspace{-10pt}
\end{table*}

\subsection{Adaptation across Different Missing Rates}

\noindent In order to further discover the efficiency of our method in helping models adapt to different missing rates, we conduct experiments with testing rates different from input rates. As shown in Table \ref{adapt_severe}, compared to ORIG method, we can see that M$^3$S significantly improves nearly all metrics by at least 1\%. It is worth noticing that a large missing rate (60\%-80\%) is adopted in the testing, and M$^3$S achieves much better performance than the other two methods. For example, the Acc-7 of M$^3$S on MOSI dataset is over 3.6\% higher than the one of ORIG+SPL-TRN method, demonstrating the capability of M$^3$S when different modalities have large missing information.

\subsection{Further Discussion and Limitations}
\noindent The qualitative results and ablation study above show that M$^3$S significantly helps baseline models improve their performance on inputs with various missing rates. However, when we apply M$^3$S to Self-MM model and conduct experiments on CMU-MOSI dataset, we find that the results show little difference from the original model's result. Besides, from Table \ref{results} we know that M$^3$S improves Self-MM's performance on SIMS dataset significantly. Hence we assume that this is because Self-MM model has good adaptability to CMU-MOSI dataset but not SIMS dataset when both datasets have a mixture of missing across modalities. Therefore, 
some models may show adaptivity to certain datasets. And M$^3$S may not significantly improve the model's performance on those datasets that model is already quite adaptive to. 

Also, as shown in Table \ref{miss_rate}, it's revealed that when inputs have a large missing rate (60\%-80\%), M$^3$S becomes limited in improving evaluation metrics. We attribute this to the change of sampling range. That is, when inputs have missing rates no more than 60\%, we can create sufficient augmented missing data to perform M$^3$S. However, when inputs have large missing rates, we can only get augmented data with missing rates restricted to a smaller range. Thus we get a smaller sampling range containing large missing rate data, which makes M$^3$S limited. 

But in general, M$^3$S method is recommended as it is easy to be added on different models and efficient in improving models' performance on multimodal sentiment analysis tasks most of the time, especially when input data has a medium missing rate. As shown in Table \ref{significance_test}, nearly all evaluation metrics' $P$-value is smaller than 0.05 in the significance test, indicating significant improvement when M$^3$S is applied.

% \begin{figure*}[t]
% 		\centering
% 		\subfigure[Orig']{
% 			\includegraphics[width = 0.3\linewidth]{figures/Orig_60.png}
% 			\label{uar}	
% 		}
%         % \quad
% 		\subfigure[Orig'Samp]{
% 			\includegraphics[width = 0.3\linewidth]{figures/Samp_60.png}
% 			\label{f1}	
% 		}
% 		% \quad
% 		\subfigure[Orig'M3S]{
% 			\includegraphics[width = 0.3\linewidth]{figures/M3S_60.png}
% 			\label{f1}	
% 		}
% 		\caption{TSNE figure of MMIN model at epoch 60.}
% \end{figure*}

\section{Conclusion and Future Work}

\noindent In this paper, we focus on a challenging problem, \textit{i.e.}, multimodal sentiment analysis on a mixture of missing across modalities, which was seldom studied in the past.
% Since few study has been developed for a mixture of missing across input modalities, 
We propose a simple yet effective method called M$^3$S to handle the problem. M$^3$S is a meta-sampling training method that follows the MAML framework and combines the sampling strategy for augmented transforms. M$^3$S maintains the advantages of meta-learning and helps SOTA models achieve superior performance on various missing input modalities.

In the experiments, we show that our method M$^3$S improves four baselines' performance and helps them adapt to inputs with various missing rates. Furthermore, M$^3$S is easy to realize in different multimodal sentiment analysis models. 
In future work, we plan to investigate how to better combine M$^3$S with other training methods and extend the method to other multimodal learning tasks.
% As mentioned in the discussion part, future work would focus on investigating the conditions of models and inputs where M$^3$S shows superior or trivial results. Also, how to better combine M$^3$S with other training methods will be investigated.

\section*{Ethical Considerations}

\noindent Our proposed method aims to help improve the performance of different SOTA methods on data with various missing rates. All experiments we conduct are based on the open public datasets (Section \ref{section:datasets}) and pretraining baseline methods (Section \ref{sec:base model}). When applying our method in experiments, there is minimal risk of privacy leakage. Furthermore, since our method is an add-on component for different baselines, it is safe to apply it as long as the baseline model provides adequate protection for privacy.

\section*{Acknowledgements}

\noindent This work is supported by National Natural Science Foundation of China (62106219). We would like to thank the anonymous reviewers for their valuable and detailed suggestions.

\bibliographystyle{acl_natbib}
\bibliography{custom}

% \appendix

% \section{Supplementary Materials}
% \label{sec:supplementary}

%     \begin{figure}[h]
% 		\centering
% % 		\setlength{\belowcaptionskip}{-1cm}
%         \includegraphics[width = 1.0\linewidth]{figures/missing_trans.png}
% 		\caption{Illustration of Augmented Missing Modality Transform}
% 		\label{missing_trans}
%     \end{figure}

% \begin{figure}[h]
% 		\centering
% % 		\subfigure[ORIG (10)]{
% % 			\includegraphics[width = 0.3\linewidth]{figures/Orig_epoch_10.png}
% % 			\label{orig10}	
% % 		}
% %          \quad
% % 		\subfigure[ORIG + SPL-TRN (10)]{
% % 			\includegraphics[width = 0.3\linewidth]{figures/Samp_epoch_10.png}
% % 			\label{origspl10}	
% % 		}
% % 		 \quad
% 		\subfigure[ORIG+M$^3$S (10)]{
% 		    \includegraphics[width = 0.46\linewidth]{figures/M3S_epoch_10.png}
% 			\label{origm3s10}	
% 		}
% % 		\quad
% % 		\subfigure[ORIG (60)]{
% % 			\includegraphics[width = 0.3\linewidth]{figures/Orig_60.png}
% % 			\label{orig60}	
% % 		}
% %          \quad
% % 		\subfigure[ORIG + SPL-TRN (60)]{
% % 			\includegraphics[width = 0.3\linewidth]{figures/Samp_60.png}
% % 			\label{origspl60}
% % 		}
% % 		 \quad
% 		\subfigure[ORIG+M$^3$S (60)]{
% 			\includegraphics[width = 0.46\linewidth]{figures/M3S_60.png}
% 			\label{origm3s60}	
% 		}
% 		\caption{TSNE Figures of MMIN at Epoch 10 and 60. Different colors represent different emotion labels.}
% \end{figure}

\end{document}